\documentclass[Afour,sageh,times]{sagej}

\usepackage{moreverb,url}
\usepackage{enumitem}
\usepackage{amsmath}

\usepackage[colorlinks,bookmarksopen,bookmarksnumbered,citecolor=red,urlcolor=red]{hyperref}

\usepackage{multirow}

\usepackage{pifont}
\usepackage[dvipsnames]{xcolor}

\newcommand{\cmark}{\ding{51}}%
\newcommand{\xmark}{\ding{55}}%
\newcommand{\rxmark}{\textcolor{red}{\xmark}}
\newcommand{\gcmark}{\textcolor{green}{\cmark}}

\usepackage[nolist]{acronym} 
\newacro{taxonomy}[T-DOM]{Taxonomy for Deformable Object Manipulation}

\usepackage{xcolor,colortbl}
\definecolor{Gray}{gray}{0.9}
\newcommand*\rot{\rotatebox{61}}

\newcommand{\PreserveBackslash}[1]{\let\temp=\\#1\let\\=\temp}
\newcommand\rcol{\rowcolor{Gray}\cellcolor{White}}

\graphicspath{{img_tex/}}

\usepackage{graphicx}
\usepackage{subfig}

\newcommand\BibTeX{{\rmfamily B\kern-.05em \textsc{i\kern-.025em b}\kern-.08em
T\kern-.1667em\lower.7ex\hbox{E}\kern-.125emX}}

\def\volumeyear{2024}

\begin{document}

\runninghead{Blanco-Mulero \textit{et al.}}

\title{T-DOM: A Taxonomy for Robotic Manipulation of Deformable Objects
}

\author{David Blanco-Mulero\affilnum{1}, Yifei Dong\affilnum{2}, Julia Borras\affilnum{1},  Florian T. Pokorny\affilnum{2} and  Carme Torras\affilnum{1}}

\affiliation{\affilnum{1}Institut de Robòtica i Informàtica Industrial, CSIC-UPC, Barcelona, Spain\\
\affilnum{2}Division of Robotics, Perception and Learning (RPL), School of Electrical Engineering and Computer Science, KTH Royal Institute of Technology, Stockholm, Sweden}

\corrauth{Institut de Robòtica i Informàtica Industrial, CSIC-UPC, Barcelona, Spain.}

\email{david.blanco.mulero@upc.edu}

\begin{abstract}
Robotic grasp and manipulation taxonomies, inspired by observing human manipulation strategies, can provide key guidance for tasks ranging from robotic gripper design to the development of manipulation algorithms.
The existing grasp and manipulation taxonomies, however, often assume object rigidity, which limits their ability to reason about the complex interactions in the robotic manipulation of deformable objects.
Hence, to assist in tasks involving deformable objects, taxonomies need to capture more comprehensively the interactions inherent in deformable object manipulation.
To this end, we introduce \acs{taxonomy}, a taxonomy that analyses key aspects involved in the manipulation of deformable objects, such as robot motion, forces, prehensile and non-prehensile interactions and, for the first time, a detailed classification of object deformations.
To evaluate \acs{taxonomy}, we curate a dataset of ten tasks involving a variety of deformable objects, such as garments, ropes, and surgical gloves, as well as diverse types of deformations.
We analyse the proposed tasks comparing the \acs{taxonomy} taxonomy with previous well established manipulation taxonomies.
Our analysis demonstrates that  \acs{taxonomy} can effectively distinguish between manipulation skills that were not identified in other taxonomies, across different deformable objects and manipulation actions, offering new categories to characterize a skill.
The proposed taxonomy significantly extends past work, providing a more fine-grained classification that can be used to describe the robotic manipulation of deformable objects.
This work establishes a foundation for advancing deformable object manipulation, bridging theoretical understanding and practical implementation in robotic systems.
\end{abstract}


\maketitle

\section{Introduction}
Robotic grasping and manipulation is a complex field due to the diverse challenges it encompasses, ranging from high-level scene understanding to the contact-rich interactions required for precise manipulation tasks.
In this context, taxonomies have been used to provide structure and understanding of the problem.
Examples range from the classic grasp taxonomies \citep{cutkosky1989grasp,feix2015grasp} that are useful for gripper design and grasp choice decision-making to more recent manipulation taxonomies \citep{bullock2012hand,paulius_2020_manipulation_tax_embedding} that define the type of interactions that can occur between the environment, the objects and the grasping agent, or the type of skills and tasks that exist \citep{worgotter2013simple}. Taxonomies and ontologies provide structure to complex problems, facilitating high-level decision-making, the development of more specialised low-level skills and the standardisation and benchmarking of the problem.

Existing taxonomies for manipulation have addressed various factors, including relative motion at contact points and between the object and environment \citep{bullock2012hand}, the use of single or multiple hands \citep{krebs_2022_bimanual_taxonomy}, the role of external forces in in-hand manipulation \citep{dafle2014extrinsic}, and broader concepts of rigid-soft interactions \citep{paulius_2020_manipulation_tax_embedding}.
However, these taxonomies have mainly focused on classifying key aspects relevant to the manipulation of rigid objects, overlooking how the object deforms when manipulated.
In this work, we address this gap by incorporating an analysis of object deformation, which can facilitate the classification of diverse manipulation skills required for handling deformable objects. 

The manipulation of deformable objects is of significant importance due to the ubiquitous presence of objects such as clothes, cables and food items in both everyday interaction scenarios and in industrial settings. Deformable object manipulation methodologies are however currently less developed compared to rigid object manipulation and pose significant challenges due to the high dimensionality of the object's configuration space. 
In order to manipulate these objects, the steps required to accomplish a task are highly influenced by the state of deformation, which may change with each interaction.
While recent reviews of deformable object manipulation comprehensively cover the progress and ongoing challenges \citep{yin_2021_modeling_learning_perception_control_for_do_manipulation, zhu2022challenges, longhini2024unfolding}, few works have tried to classify in a systematic and descriptive way how the deformation affects the 
planning and execution of each manipulation step.
One of the key limitations to identify the deformation of an object comes from the complexity of accurately measuring its state~\citep{ sanchez_2018_manipulation_sensing_do_survey, yin_2021_modeling_learning_perception_control_for_do_manipulation, zhu2022challenges}, which involves the partial observability of the object.
Existing approaches address this by using quantitative metrics that measure the deformation of objects such as ropes~\citep{liu_2023_robotic_do_manip_differentiable_compliant_pbd} or clothes~\citep{coltraro2023representation}, and qualitative metrics or labels that describe its state, e.g. identifying crossings in ropes~\citep{huang_2024_untangle_dlos_segmentation}, or the folded state of cloths~\citep{jimenez_2017_overview_cloth_state_recognition, garcia_2022_knowledge_representation_cloth}.
Due to the complexity of mapping diverse deformation modes to a quantitative metric, to build a taxonomy that can describe the deformation, this work adopts a qualitative classification that incorporates the forces in the manipulation task that produce the deformation.

In order to establish a framework for understanding robotic manipulation of deformable objects, this work proposes the \ac{taxonomy}, a taxonomy encompassing information about the type of deformation of the object, the robot manipulator motion, the end-effector grasp, as well as the interactions of the manipulated deformable object with the environment or the end-effector.
The taxonomy is evaluated using a dataset with varied manipulation actions and deformable objects. 
The analysis reveals that the proposed categories allow clear differentiation of manipulation strategies, even when the same object is subjected to varying deformations.
Furthermore, our work discusses several applications for the proposed taxonomy including gripper design and learning of manipulation skills, along with future research directions.

Our main contributions can be summarised as follows:
\begin{itemize}
    \item \acs{taxonomy}: a taxonomy that qualitatively categorises deformation, motion and interactions in robotic manipulation of deformable objects, allowing for a structured analysis of these tasks.
    \item A qualitative classification of bending deformation based on the structure of the object state, distinguishing between structured and unstructured bending to better capture the type of deformation.
    \item A dataset of 10 deformable object manipulation tasks, including garments, silicone meat phantoms, bags, and surgical deformable objects.
    \item An analysis of the taxonomy, showcasing its effectiveness in differentiating manipulation skills across tasks in the dataset and evaluating the critical role of deformation for this classification.    
\end{itemize}

\section{Related Works}\label{sec:related-works}
This section starts by providing a review and discussion of existing grasping and manipulation taxonomies, highlighting the gaps in the literature, followed by a short review of deformable object manipulation works and how these address the deformation of the object.

\subsection{Manipulation Taxonomies}
In the field of manipulation, several taxonomies have been proposed to classify diverse aspects of grasping and manipulation.
\cite{cutkosky1989grasp} developed one of the earliest grasp taxonomies, focusing on anthropomorphic grasp types for manufacturing tasks. This taxonomy classifies grasps based on the geometric and functional characteristics of the hand and the object, including criteria such as prehensility and grasp dexterity. 
\cite{bullock2012hand} expanded on this by introducing a hand-centric taxonomy that addresses both human and robotic manipulation. 
This taxonomy offers a detailed analysis of the contacts and motions involved in manipulation tasks, including characteristics such as the presence of motion at the point of contact.
More recently, \cite{krebs_2022_bimanual_taxonomy} proposed a taxonomy for bimanual manipulation, addressing the complexity of using two hands simultaneously. This classification takes into account the coordination of the hands as well as spatial and temporal constraints.
Nevertheless, these taxonomies assume that the manipulated object is rigid, and do not classify interactions with the object that may result in its deformation.

Few recent works have proposed classifications that also consider the manipulation of deformable objects.
\cite{borras_2020_grasp_cloth_analysis} provided an analysis of the types of grasps used in cloth manipulation, considering the number of contacts and the geometries of the manipulation agents.
However, this classification is restricted to grasps of textile materials, disregarding other aspects of manipulation such as the motion of the end-effector or the deformation of the object.
\cite{paulius_2020_manipulation_tax_embedding} introduced a more generic motion taxonomy that takes into account the deformation of the manipulated object by classifying the engagement type as rigid or soft, and the structural outcome of the object that, if deformed, could exhibit a temporal or permanent deformation.

Although the aforementioned taxonomies have provided comprehensive classifications of grasping and manipulation tasks, these have not incorporated important aspects relevant to the manipulation of deformable objects, such as the type of deformation.
Our proposed taxonomy addresses these gaps by classifying the type of deformation based on the forces involved in deforming these objects.
In addition, our work categorises the robot motion during the manipulation, as well as the interactions between objects and their environment, which are critical in real-world manipulation tasks involving deformable objects.

\begin{figure*}
\centering
\def\svgwidth{\linewidth}
{
    \fontsize{9}{9}
\begingroup%
  \makeatletter%
  \providecommand\color[2][]{%
    \errmessage{(Inkscape) Color is used for the text in Inkscape, but the package 'color.sty' is not loaded}%
    \renewcommand\color[2][]{}%
  }%
  \providecommand\transparent[1]{%
    \errmessage{(Inkscape) Transparency is used (non-zero) for the text in Inkscape, but the package 'transparent.sty' is not loaded}%
    \renewcommand\transparent[1]{}%
  }%
  \providecommand\rotatebox[2]{#2}%
  \newcommand*\fsize{\dimexpr\f@size pt\relax}%
  \newcommand*\lineheight[1]{\fontsize{\fsize}{#1\fsize}\selectfont}%
  \ifx\svgwidth\undefined%
    \setlength{\unitlength}{606.1534395bp}%
    \ifx\svgscale\undefined%
      \relax%
    \else%
      \setlength{\unitlength}{\unitlength * \real{\svgscale}}%
    \fi%
  \else%
    \setlength{\unitlength}{\svgwidth}%
  \fi%
  \global\let\svgwidth\undefined%
  \global\let\svgscale\undefined%
  \makeatother%
  \begin{picture}(1,0.31547749)%
    \lineheight{1}%
    \setlength\tabcolsep{0pt}%
    \put(0.4773209,0.01185449){\color[rgb]{0,0,0}\makebox(0,0)[lt]{\lineheight{1.25}\smash{\begin{tabular}[t]{l}Bending\end{tabular}}}}%
    \put(0,0){\includegraphics[width=\unitlength,page=1]{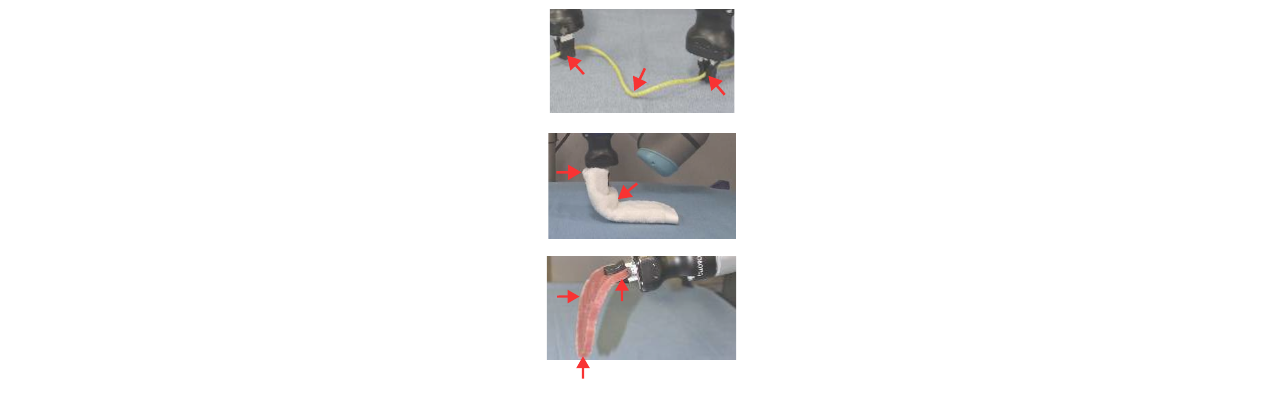}}%
    \put(0.29346747,0.0099357){\color[rgb]{0,0,0}\makebox(0,0)[lt]{\lineheight{1.25}\smash{\begin{tabular}[t]{l}Tension\end{tabular}}}}%
    \put(0,0){\includegraphics[width=\unitlength,page=2]{types_of_deformation_v6.pdf}}%
    \put(0.06050316,0.01137479){\color[rgb]{0,0,0}\makebox(0,0)[lt]{\lineheight{1.25}\smash{\begin{tabular}[t]{l}Compression\end{tabular}}}}%
    \put(0,0){\includegraphics[width=\unitlength,page=3]{types_of_deformation_v6.pdf}}%
    \put(0.66993586,0.0099357){\color[rgb]{0,0,0}\makebox(0,0)[lt]{\lineheight{1.25}\smash{\begin{tabular}[t]{l}Torsion\end{tabular}}}}%
    \put(0,0){\includegraphics[width=\unitlength,page=4]{types_of_deformation_v6.pdf}}%
    \put(0.8773967,0.0099357){\color[rgb]{0,0,0}\makebox(0,0)[lt]{\lineheight{1.25}\smash{\begin{tabular}[t]{l}Shear\end{tabular}}}}%
    \put(0,0){\includegraphics[width=\unitlength,page=5]{types_of_deformation_v6.pdf}}%
  \end{picture}%
\endgroup%
}
    \caption{Robotic manipulation of multiple deformable objects showing compression, tension, bending, torsion and shear deformation. Shear deformation examples include a cloth shear deformation from the open-source dataset provided by~\cite{wang_2011_arcsim_cloth}, and a sponge manipulated by a human.
    } 
\label{fig:deformation}
\end{figure*}

\subsection{Types of Deformation Under Manipulation}

The robotic manipulation of deformable objects involves different types of deformation, such as tension, compression or bending.
For instance, \cite{saha2007manipulation} focused on the manipulation of deformable linear objects by applying concepts of knot theory to manipulate cables and plan knotting strategies.
More recently, \cite{luo2024multi} provided insights into efficient routing strategies of linear deformable cables through bending.
In garment manipulation, \cite{vandenBerg2011_gfold} focused on sequentially folding cloths by identifying the structure of the bending deformation, denoted as g-folds.
\cite{li2015folding} identified parameters such as the shear resistance of clothes for planning their manipulation by measuring the bending deformation of the object.
\cite{longhini2021textile} identified textile properties based on their mechanical responses through tension and twisting applied by a dual arm. 
For manipulating volumetric deformable objects, \cite{shen2022acid} explored the tension, bending, and compression of plush toys. 
\cite{thach2022learning} focused on shape control of volumetric deformable objects through bending, developing control strategies from limited visual input. 

This paper presents a classification and analysis of the aforementioned types of deformation commonly observed in tasks studied within deformable object manipulation research.

\section{Properties of Deformable Object Manipulation}\label{sec:properties}
This section introduces three key aspects of manipulating a deformable object that serve as the foundation for developing the proposed taxonomy: the concept and types of deformation, the motion during manipulation and its relationship to energy changes of the manipulated object, and the interactions between the manipulated object and its surroundings.

\subsection{Deformation Concepts and Terminology}\label{sec:properties-deformation}
The term \textit{deformation} is extensively used in the context of robotic manipulation of deformable objects, encompassing various aspects such as perception, modelling, and control~\citep{Cheong_2019_modeling_deformables_robot_manipulation_review, arriola_2020_modeling_do_robotic_manipulation_review, yin_2021_modeling_learning_perception_control_for_do_manipulation}.
However, throughout these works the concept of deformation is ambiguous and left to the reader to conceptualise.

The Dictionary of Engineering defines deformation as ``\textit{any alteration of shape or dimensions of a body caused by stresses, thermal expansion or contraction, chemical or metallurgical transformations, or shrinkage and expansions due to moisture change}''~\citep{mcgraw2002dictionary}.
This definition opens a wide spectrum involving thermal, chemical, and metallurgical transformations.
These transformations can often be neglected when performing robotic manipulation of deformable objects by assuming constant environmental conditions. 
\cite{sanchez_2018_manipulation_sensing_do_survey} state that "a deformation occurs when an external force applied to an object results in the object changing its shape".
More generally, deformation refers to any change in the shape of an object. 
For a known object, it is considered deformed if its shape differs from a predefined canonical shape~\citep{Chi_2021_garmentnets, canberk_2022_clothfunnels}.
Conversely, if the object retains its canonical shape, it is considered as not deformed. 
However, in the absence of prior knowledge of the object's shape, deformation is identified by any deviation from an initially observed object shape.

In this work, we assume that the forces that deform an object in a robotic manipulation task can originate from: 1) a robot manipulator, 2) a tool manipulated by a robot, or 3) objects in the environment interacting with the manipulated object.
Considering the forces that a robot applies during grasping, or while performing a motion to complete a task, it seems intuitive to relate these forces to the object deformation they can induce.
Following this, deformation can be characterised by the direction of the applied forces (see Figure~\ref{fig:deformation}). The type of deformation can be classified as compression, tension, bending, torsion, and shear~\citep{callister2020_materials_science}.
A detailed definition of these deformations is provided below, accompanied by examples of common tasks in robotic manipulation tasks.

\begin{figure*}
\centering
\def\svgwidth{\linewidth}
{
    \fontsize{10}{9}
    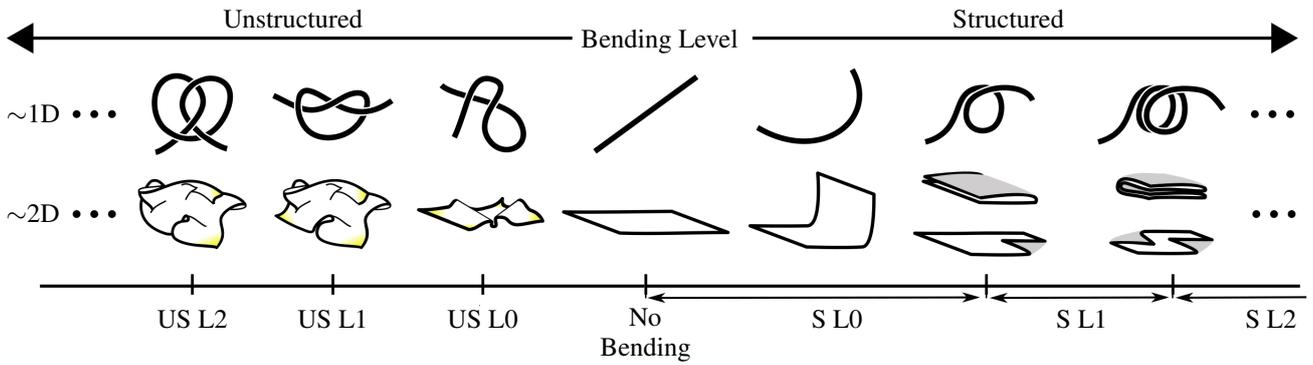}
    \caption{Qualitative classification levels for bending deformation as structured or unstructured for 1D and 2D deformable objects. The structured level is classified by loops and \textit{g-folds}~\citep{vandenBerg2011_gfold}, for 1D and 2D objects, respectively.
    The unstructured level is classified by knots for 1D objects, and as the number of accessible corners for a cloth flattening task. 
    }
    
    \label{fig:bending_example}
\end{figure*}

\paragraph{Compression.} This deformation results from applying two opposite forces towards the object along a specific axis, causing the object to shorten in the direction of the applied forces.
This deformation can also occur when applying a single force to an object placed on a surface in the direction perpendicular to the surface.
This type of deformation is common when shaping volumetric deformable objects~\citep{shi_2024_robocraft_gns_ijjr}.
As pointed out by~\cite{sanchez_2018_manipulation_sensing_do_survey}, 1D and 2D deformable objects such as ropes and textiles are characterised as objects without compression stress. 
This follows from the fact that these objects are usually modelled following an incompressibility assumption~\citep{kashani_2018_shear_woven}.
However, for certain deformation configurations such as a folded or stacked 2D object, the object has significant volumetric structure and in such cases, the assumption of incompressibility can be relaxed, allowing 2D objects, such as cloth, to undergo compression deformation, e.g. grasping a folded cloth, see Figure~\ref{fig:deformation}.

\paragraph{Tension.} This deformation can be seen as the inverse of compression. Given one fixed point on the object such as a grasping point, tension is generated when a force is applied away from the object, stretching the object along the axis of such forces without tearing it.
This deformation is common when a deformable object is stretched, for example, when placing a tight cover on a chair or a bed sheet, stretching a cloth in the air for placing it flat on a table~\citep{ha2021flingbot}, and when knotting ropes~\citep{lee_2015_force_based_demo_manipulation} or plastic bags~\citep{gao_2023_knotting_bags_iterative_modeling}.

\paragraph{Bending.}
Also referred to as flexural deformation, this deformation appears when applying an external force perpendicular to the object's longitudinal axis, causing it to curve.
The Dictionary of Engineering defines bending as the deformation of points in an elastic body that were originally straight, causing them to displace and form a plane curve~\citep{mcgraw2002dictionary}.
Notably, even after the external force is removed, the bending deformation can persist due to the internal forces within the object, which may prevent it from returning to its original shape.
We refer the reader to the existing literature on deformable object modelling~\citep{Cheong_2019_modeling_deformables_robot_manipulation_review, arriola_2020_modeling_do_robotic_manipulation_review, coltraro2022inextensible}, which provides a detailed explanation of how these internal forces can be approximated using different mathematical models.
Consequently, bending is a common deformation encountered in deformable object manipulation tasks, such as shaping ropes and clothes~\citep{chi_2024_irp_deformables_ijrr}, or deforming plush toys~\cite{shen2022acid}.

\paragraph{Torsion.} A torsional or twisting deformation is generated when a torque is applied around the longitudinal axis of the manipulated object, given that the axis is fixed in one of its sides.
This can occur as well when two torques are applied in opposite directions along the longitudinal axis of the object.
Similarly to the compression deformation, deformable linear objects and cloth-like objects can be subjected to this deformation when they possess sufficient volumetric structure. 
This deformation can be seen when twisting a thread \citep{saha2006motion}, or wringing out a sponge or a towel to dry it out~\citep{bai_2016_dexterous_cloth_manipulation_sim_qp}, see Figure~\ref{fig:deformation}.

\paragraph{Shear.} A shear deformation is caused when two external forces are applied in opposite directions along two distinct parallel lines of the object~\citep{mcgraw2002dictionary}. Manipulation actions that cause a shear deformation are common when performing system identification~\citep{wang_2011_arcsim_cloth}.
In textiles, the elasticity along different pulling lines may differ dramatically due to fabrication processes. That elasticity along diagonal lines is an important parameter to consider in this particular setting. 
Shear deformation can also be observed on volumetric deformable objects, such as meat \citep{pierre2023discovering}.
Nevertheless, this deformation is less common in 1D and 2D deformable objects, where usually the plane section principle is followed and therefore shear deformation is neglected~\citep{yu_2016_1d_planesection}.

The concepts of compression, tension and torsion can be intuitively understood as squeezing, stretching and twisting an object, respectively, and can therefore be clearly identified when manipulating deformable objects.
However, while the concept of bending is widely utilised in the literature~\citep{li2015folding, shen2022acid, thach2022learning}, it does not provide direct insight of the shape or state of the object.
For instance, both a crumpled and a folded cloth exhibit bending deformation, yet their shapes are essentially different.
The challenge of defining bending deformation arises from the infinite number of degrees of freedom in deformable objects, where the internal forces can generate small localised curvature changes in the object.

Consequently, it is challenging to categorise the deformation of an object simply as bending to understand the object state.
To identify different bending deformation cases, we introduce two concepts: \textit{structured bending} and \textit{unstructured bending}, depicted in Figure~\ref{fig:bending_example}.
These types of bending deformation are characterised by the structural arrangement of the deformation.
If ordered bends exist, as happens when folding, we call the bending structured. 
On the contrary, if the bending deformation is chaotic or wrinkled, we call it unstructured.
Given the distinct nature of 1D and 2D deformable objects, we provide separate metrics to recognize the degree of structuredness.
For the 1D case, we draw inspiration from knot theory and deformable linear object manipulation~\citep{matsuno_2006_dlo_know_invariants, saha2007manipulation}, while for the 2D case, we adopt the concept of \textit{g-folds} as proposed by~\cite{vandenBerg2011_gfold}. 

\paragraph{Structured Bending.} 
To quantify the level of structure in a bending deformation we use the number of removable crossings, such as loops, or \textit{g-folds} as metrics for 1D and 2D deformable objects, respectively.
In 3D objects, the structured bending is measured in the plane curve where the bending deformation takes place.

\paragraph{Unstructured Bending.}  
The idea is to measure the level of unstructuredness by how much one needs to manipulate the object to remove the bending.
That is, displacing the points that form a plane curve to their originally straight configuration.
For 1D objects, we propose to use the linking number of knots, which gives an idea of how difficult to untie a knot is.
For 2D, the manipulation required to flatten an object is usually related to being able to grasp certain keypoints, such as the shoulders of a T-shirt or the corners/edges of a napkin. Therefore, we propose to measure unstructured bending deformation by the number of graspable or accessible keypoints (see Figure~\ref{fig:bending_example}).

Both structured and unstructured bending deformations can take place at the same time. 
For example, a rope can present loops as well as knots, and cloths can present g-folds as well as unstructured wrinkles.

\subsection{Energies during Manipulation Motion}\label{sec:properties-energy-motion}
The manipulation of a deformable object can be conceptualised as determining the configuration of exerted forces that brings the object to the desired state.
Prior works on deformable object manipulation have addressed the optimisation problem of finding the motion to reach the desired minimum-energy state or identifying the robustness of a grasp from an energy perspective~\citep{bretl_2014_elastic_rod_optimal_control_geometric, dong2023quasi, dong2024caging}.
In this context, the end-effector motion can be prescribed based on the predominant energies of the object when moved from one state to the desired one.
These energies are the kinetic and potential energy of the object, where the latter can be further divided into gravitational potential energy and elastic potential energy. 
While recent works simplify the problem of finding the minimum energy-state motion by using e.g. predefined motions that neglect the associated energies~\citep{ha2021flingbot}, these approaches share the underlying goal of moving the object into the desired state.

In this work, we relate the energies of motion to the manipulation classification by ~\cite{Mason_2001_dynamic_quasistatic_book}. 
Here, \textit{dynamic manipulation} is defined as manipulation tasks dominated by inertial forces, that is, the acceleration and mass of the object. 
Hence, the motion is dominated by the kinetic energy of the manipulated object.  
By contrast, \textit{quasi-static manipulation} is defined as tasks dominated by frictional and impact forces, neglecting inertia. 
In deformable object manipulation, tasks that require transporting the object or performing pick-and-place motions are defined as quasi-static, which are dominated by the gravitational energy.
Similarly, tasks where the success depends on the elasticity of the deformable can be defined as quasi-static, which are dominated by the elastic energy of the object.
This energy is particularly relevant in deformable objects like elastic rods, whose manipulation can be formulated as the optimisation of the elastic energy of the rod~\citep{bretl_2014_elastic_rod_optimal_control_geometric}.

In addition, Mason classifies manipulation as kinematic or static. In this work, we neglect these two categories since deformable object manipulation tasks cannot be described uniquely by the kinematics of the end-effector, and the friction of the object due to collisions and self-collision needs to be taken into account even when performing what is denoted as static manipulation.

\begin{figure*}[t]
\centering
\def\svgwidth{\linewidth}
{
    \fontsize{7}{12}
    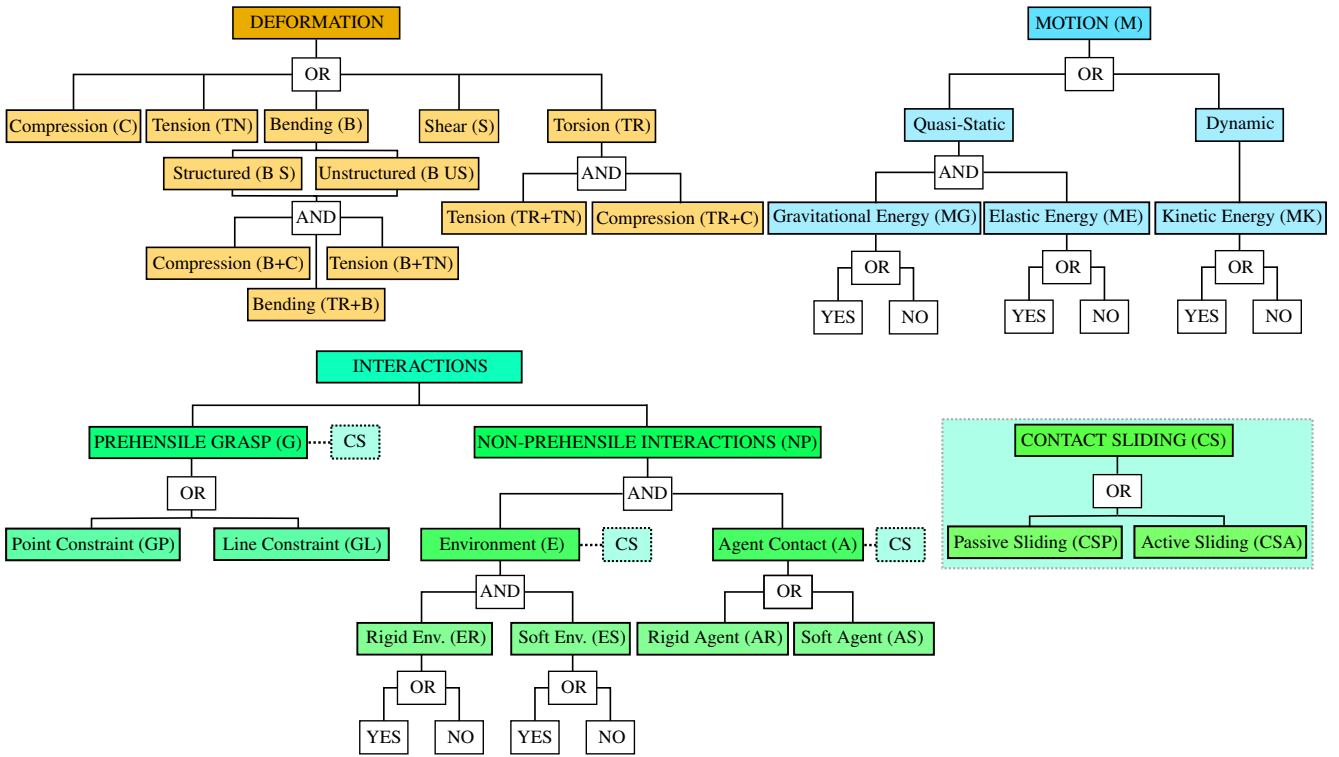}
    \caption{The proposed taxonomy is composed of deformation (D), motion (M), and interactions. The interactions are classified as prehensile grasp (G) and non-prehensile interactions (NP), and contact sliding (CS) as a special case of interaction that can take place in both prehensile grasps as well as each sub-category of non-prehensile interactions, shown by a dashed line connecting the CS block. Each sub-category is shown with its associated short tag.} 
\label{fig:taxonomy}
\end{figure*}
\subsection{Interactions}\label{sec:sec:properties-extrinsic}

Robotic manipulation is accomplished through physical interactions of the robot with the target manipulation object.
This interaction can be defined as \textit{prehensile} if the end-effector contact forces with the object can stabilise it without external forces.
By contrast, if the interaction involves forces such as gravity or a reaction force from a contact with the environment the interaction is denoted as \textit{non-prehensile}.
Grasping an object can be achieved by either prehensile or non-prehensile grasps.
Prehensile grasps for rigid objects have been extensively studied and classified in works like \cite{cutkosky1989grasp} and \cite{feix2015grasp}. \cite{cutkosky1989grasp} further classified grasps by their purpose or function, distinguishing between power and precision grasps, while \cite{feix2015grasp} categorised grasps based on the number of fingers involved in exerting the opposite forces. 
On the contrary, the diversity of prehensile grasps for deformable objects has been less actively studied, partly because a pinch grasp usually suffices to perform a wide range of deformable object manipulation tasks.
\cite{borras_2020_grasp_cloth_analysis} introduced a framework to classify textile grasps based on the geometry of the contact patches, showing that although pinch grasps are common, more complex grasps play a significant role in dynamic manipulation. This framework included both environmental and robot end-effector contacts, as well as bi-manual grasps.

Non-prehensile interactions using contacts with surfaces of the environment have been largely considered in the literature for rigid objects, such as in \citep{bullock2012hand, eppner2015exploitation, dafle2014extrinsic} and less frequently addressed for deformable objects \citep{borras_2020_grasp_cloth_analysis}. 
Additionally, research has explored combinations of prehensile grasps with environmental contacts \citep{chavan2018stable}.
Furthermore, non-prehensile grasps using acceleration forces present more challenges. These are considered in the context of extrinsic dexterity \cite{dafle2014extrinsic}, which refers to the use of any external forces to enhance manipulation.

All in all, these concepts are used to build the classification developed from Section~\ref{sec:taxonomy-grasp} to Section~\ref{sec:taxonomy-contact-sliding}.
\section{A Taxonomy for the Manipulation of Deformable Objects}\label{sec:taxonomy}
This section describes \ac{taxonomy}, a taxonomy for robotic manipulation of deformable objects, depicted in Figure~\ref{fig:taxonomy}, which includes short tags associated with each category of the classification.
The taxonomy is divided into three main categories: object deformation, robot motion, and interactions.
For clarity, Table~\ref{table:related_works} provides a comparison of the categories and sub-categories from our proposed taxonomy with those from existing grasping and manipulation taxonomies. 
The remaining sections will explain how the proposed categories compare to the related literature.

\subsection{Deformation Classification}\label{sec:taxonomy-deformation}
The types of deformation considered in \ac{taxonomy} build on those introduced in Section~\ref{sec:properties-deformation}.
In addition to compression (\textbf{C}), tension (\textbf{TN}), torsion (\textbf{TR}), and shear (\textbf{S}), we distinguish between different levels of structured bending and unstructured bending.
The first structured bending level (\textbf{S L0}) identifies the transition from no bending deformation to one crossing or g-fold.
The subsequent structured levels refer to increasing the numbers of loops or g-folds, e.g. \textbf{S L2} refers to either 2 loops or 2 g-folds.
For unstructured bending, the qualitative metrics are discrete.  As previously discussed, for 1D objects the first level (\textbf{US L0}) is defined by removable crossings, and for 2D objects by almost flat clothes with wrinkles, where all keypoints are visible. Further levels of unstructured follow Fig.~\ref{fig:bending_example}
.

Furthermore, we also address several combinations of these deformations, which often occur simultaneously during the manipulation of deformable objects. Below are a few examples:

\begin{itemize}
    \item \textbf{C+B: Compression \& Bending.} In the 2D-compression example shown in Figure~\ref{fig:deformation}, the towel is folded (bending) and the grasp performed to pick it up compresses it. 
    \item \textbf{TN+B: Tension \& Bending}. For instance when folding a piece of cloth while applying tension at the grasped points to prevent wrinkles \citep{lee_2015_force_based_demo_manipulation}. 
    \item \textbf{TN+TR: Tension \& Torsion}. In Figure~\ref{fig:deformation} torsion, a twisted towel undergoes torsional deformation as it is wrung by a bimanual system.
    Simultaneously, the robots apply tension by pulling the towel in opposite directions.
\end{itemize}

As discussed in Section~\ref{sec:related-works}, this classification provides a more comprehensive understanding of object deformation than prior taxonomies, which either overlook deformable objects or fail to account for the forces driving the deformation, limiting their ability to differentiate between different types of deformation, see Table~\ref{table:related_works}.  
Note that we leave for future work incorporating into the taxonomy deformations that break the object or deform it in a permanent way, such as plastic deformations, which have been less actively explored as discussed in Section~\ref{sec:discussion-def-categories}.

\begin{table*}[h]
\small\sf\centering
\caption{Comparison of categories and sub-categories covered by grasping and manipulation taxonomies found in literature and the proposed taxonomy. The categories are divided into distinctions of bimanual manipulation, robot motion, prehensile grasp, interactions and deformation of the manipulated object.  }\label{table:related_works}
\begin{tabular}{cc c c c c c c}
\rot{Category} & \rot{Sub-category} & \rot{\parbox{2.9cm}{\cite{cutkosky1989grasp}\\ \cite{feix2015grasp}}}
& \rot{\cite{bullock2012hand}} &
\rot{\cite{borras_2020_grasp_cloth_analysis}} & \rot{\cite{paulius_2020_manipulation_tax_embedding}} & \rot{\cite{krebs_2022_bimanual_taxonomy}}
& \rot{\textbf{Ours}} \\
\midrule
 \multicolumn{2}{c}{Bimanual Manipulation} & \rxmark & \gcmark & \gcmark & \gcmark & \gcmark & \gcmark \\
 \midrule
 \multirow{2}{1.3cm}{Robot Motion} & Trajectory & \rxmark & \gcmark & \rxmark & \gcmark  & \rxmark & \rxmark  \\
 & Energy & \rxmark & \rxmark & \rxmark & \rxmark  & \rxmark & \gcmark\\
 \midrule
\multirow{3}{1.3cm}{Prehensile\\ Grasp} & Anthropomorphic & \gcmark & \gcmark  & \rxmark & \rxmark & \gcmark & \rxmark \\
 & Nr. Contacts & \gcmark & \rxmark & \gcmark & \rxmark & \rxmark & \gcmark \\
 & Robotic & \rxmark & \gcmark & \gcmark & \rxmark & \rxmark & \gcmark \\
 \midrule
  \multirow{4}{1.3cm}{Interactions} & Motion at Contact & \rxmark & \gcmark & \rxmark & \rxmark & \rxmark & \gcmark  \\
  & Non-Prehensile & \rxmark & \gcmark & \gcmark & \rxmark & \rxmark & \gcmark  \\
 & Rigid/Soft & \rxmark & \rxmark & \rxmark & \gcmark & \rxmark & \gcmark \\
& Contact Duration & \rxmark & \rxmark & \rxmark & \gcmark & \rxmark & \rxmark \\
\midrule
\multirow{3}{1.3cm}{Deformable\\Objects} & Objects & \rxmark & \rxmark & \gcmark  & \gcmark  & \rxmark & \gcmark    \\
& Elastic/Plastic & \rxmark & \rxmark & \rxmark & \gcmark & \rxmark & \rxmark \\
 & Force Direction & \rxmark & \rxmark & \rxmark & \rxmark & \rxmark & \gcmark  \\
\bottomrule
\end{tabular}
\end{table*}

\subsection{Motion Classification}\label{sec:taxonomy-motion}
For the proposed taxonomy, following the definitions outlined in Section \ref{sec:properties-energy-motion}, we introduce the following classification:

\begin{itemize}
    \item \textbf{MK: Motion $\rightarrow$ Kinetic}. This includes dynamic motions as described in the previous section, dominated by inertial forces. Some examples include swinging a cloth to unfold it~\citep{ha2021flingbot}, or hitting a target with a rope~\citep{chi_2024_irp_deformables_ijrr}.
    \item \textbf{ME: Motion $\rightarrow$ Elastic}
    This includes quasi-static manipulations where the task success depends mostly on the elastic forces of the object. Examples include shaping a rope~\citep{bretl_2014_elastic_rod_optimal_control_geometric} or a volumetric object~\citep{shi_2024_robocraft_gns_ijjr}, and securing a mask string~\citep{dong2023quasi}.
    \item \textbf{MG: Motion  $\rightarrow$ Gravitational}.
    This tag includes quasi-static manipulations where the elasticity of the object can be neglected. Examples include transporting a cloth to store it~\citep{yang2024equibot} or motions performed to approach the object.
    \item \textbf{MGE: Motion $\rightarrow$ Gravitational-Elastic}.
    These are cases where both elastic and gravitational energies play a role in deformable object manipulation.
    For instance, \cite{dong2023quasi} present a scenario where a fish is lifted from a table using a scoop. While the scoop containing the fish is lifted up, the fish is subject to elastic force due to the deformation of its body under gravity, accompanied by a change of gravitational energy. 

\end{itemize}

Compared with the taxonomy definition (\textit{motion} and \textit{no motion}) in \cite{bullock2012hand}, we further sub-categorise \textit{motion} in finer granularity.
By contrast, our classification is less detailed than the one proposed by~\cite{paulius_2020_manipulation_tax_embedding}, which classifies the different prismatic and revolute axes of the motion trajectory.

\subsection{Prehensile Interaction Classification}\label{sec:taxonomy-grasp}

Building on the definitions presented in Section~\ref{sec:sec:properties-extrinsic}, we classify interactions into prehensile grasps and non-prehensile interactions.
In this work, prehensile grasps are defined using a variation of the prehension geometry framework from \cite{borras_2020_grasp_cloth_analysis}. Rather than focusing on the geometry of the contact patches that generate the opposing forces defining the grasp, we categorise grasps based on the geometry they constrain on the object.
Hence, the classification proposed for prehensile grasps is as follows:
\begin{itemize}
    \item \textbf{GP: Grasp $\rightarrow$ Point.} Pinch grasps are a clear example of grasps that constrain a point, where the position of a single point on the object is controlled.
    \item  \textbf{GL: Grasp $\rightarrow$ Line.} Constraining the position of a line of an object can be used to perform tasks such as cloth folding more efficiently~\citep{sugiura2010foldy} or to fold in the air \citep{borras_2020_grasp_cloth_analysis}.
\end{itemize}
These categories can be further extended to more complex geometries as in the taxonomy by \citep{borras_2020_grasp_cloth_analysis}, which is the only prior taxonomy that does take into account the number of contacts in prehensile interactions of deformable objects.
However, for simplicity, our classification is restricted to the geometries that can be constrained with parallel jaw grippers, which are more common in deformable object manipulation tasks.

\subsection{Non-Prehensile Interactions Classification}\label{sec:taxonomy-non-prehensile}
In this work, non-prehensile interactions are classified into contacts of the manipulated object with the \textit{environment} or with the \textit{agent} that manipulates the object.
Furthermore, we distinguish between interactions made by a soft or rigid object, similar to \cite{paulius_2020_manipulation_tax_embedding}.

\paragraph{Non-Prehensile Agent Contact}
This category encompasses interactions performed by the robot hand by applying external forces. Specifically, one of the opposing forces is applied by an agent (robot or tool), while the other force may originate from the environment or gravity.
Examples of this category include basket grasps~\citep{shirizly2024selection}, a caging grasp~\citep{diankov2008manipulation}, or pressing an end-effector against a table for flattening a cloth. In this category, we consider that the agent that creates the contact can be either a robot or a tool held by a robot, where the tool acts as the end-effector, which is actively controlled by the robot \citep{vahrenkamp2012manipulability}.  Here, we follow the assumption that we do not use tools that can perform a prehensile grasp, hence, tool contacts are classified as part of the category of non-prehensile contacts.

The agent contacts are classified into rigid or soft as:

\begin{itemize}
    \item \textbf{N-P.AR: Non-Prehensile $\rightarrow$ Agent Contact $\rightarrow$ Rigid}.  
    Examples of this category are tools for manipulating deformable objects,  such as using a shovel \citep{dong2023quasi} or a rigid robot hand.
    \item \textbf{N-P.AS: Non-Prehensile $\rightarrow$ Agent Contact $\rightarrow$ Soft}.
    Some examples include holding a plastic bag to insert objects~\citep{chen_2023_autobag} or utilising a soft robot hand.
\end{itemize}

\paragraph{Non-Prehensile Environment}
This category refers to grasps where both opposing forces are exerted by extrinsic forces, which may be gravity or contacts with the environment.
This category could be categorised as no-interaction, but considering environment contacts enables us to further distinguish the interactions that lead to the deformation of an object.
The non-prehensile environment interactions are classified as follows:
\begin{itemize}
    \item \textbf{N-P.ER: Non-Prehensile $\rightarrow$ Environment $\rightarrow$ Rigid}. This includes interactions with rigid surfaces, such as an object resting on a table.
    \item \textbf{N-P.ES: Non-Prehensile $\rightarrow$ Environment $\rightarrow$ Soft}. Examples of this category are interactions of the deformable object with other deformable objects, e.g. a towel placed on a pile of towels.
    \item \textbf{N-P.ERS: Non-Prehensile $\rightarrow$ Environment $\rightarrow$ Rigid \& Soft}.
    Both types of environment interactions—rigid and soft—can occur simultaneously.
    For instance, a set of grapes partially wrapped in an elastic protective cover lying on a hard tray surface demonstrates both rigid and soft environment contacts.
\end{itemize}

\subsection{Contact Sliding}\label{sec:taxonomy-contact-sliding}
Along with the aforementioned interactions, our taxonomy classifies one special case that is notably relevant in deformable object manipulation, which is contact sliding.
It is conceptually equivalent to the category of motion at contact discussed by~\cite{bullock2012hand} for rigid objects. 
This interaction can be seen when sliding a robot end-effector on a cloth to find its edge e.g. unfolding the garment~\citep{proesmans_2023_tactile_manipulation_cloth_unfolding}, or sliding through a cable for its posterior insertion~\citep{yu_2021_cable_manip_reactive_sliding}. 
In cases where motion at contact exists, the contact point on the robot end-effector usually remains unchanged, but the points on the object translate or rotate with respect to a fixed contact frame on the robot.
In deformable object manipulation, we primarily consider translational motion at contact, i.e. sliding at contact, although a rotation could also be possible, and appears in rigid object in-hand manipulation as pivoting.
Sliding is a characteristic of every contact, therefore, it is a branch of the taxonomy that connects to both prehensile grasping, environment and agent contact interactions, see Figure~\ref{fig:taxonomy}.
The category of contact sliding is classified as follows:
\begin{itemize}
    \item \textbf{CSA: Contact Sliding $\rightarrow$ Active}: Active sliding takes place when manipulated object remains static while the end-effector moves, modifying its contact points with the object without grasping it.  Some examples include and end-effector grasping and slidinf along an edge of a cloth \cite{kondo2022development} or the tip of an end-effector sliding along the surface of a cloth to flatten it. 
    \item \textbf{CSP: Contact Sliding $\rightarrow$ Passive}: Passive sliding occurs when the end-effector remains static at the contact and the object slides pulled by an external force, such as another robot arm or gravity. An example could be an object falling from a platform grasp. 
\end{itemize}

\subsection{Bimanual manipulations}

All categories describing end-effector actions such as motion, grasps, or active contacts- refer to a single robot arm.
However, many deformable object manipulation tasks can be performed more effectively using a bimanual system~\citep{ha2021flingbot, zhang_2022_assisted_dressing_science, almaghout_2024_comanip_dlos}.
Rather than creating duplicated categories in \ac{taxonomy}, as in \cite{borras_2020_grasp_cloth_analysis, krebs_2022_bimanual_taxonomy}, we propose to decouple the tags for each arm involved in the task, e.g. left and right arm. 
Further details about this decoupling are provided in Section~\ref{sec:analysis-task-segmentation}.

\begin{figure*}[t]
\centering
\subfloat[Task 1: Fold Towel. \ac{taxonomy} no-zero constant category (omitted) is NP Env (R).\label{fig:task-transitions-1}]{
    \def\svgwidth{0.8\linewidth}
    \fontsize{7}{7}
    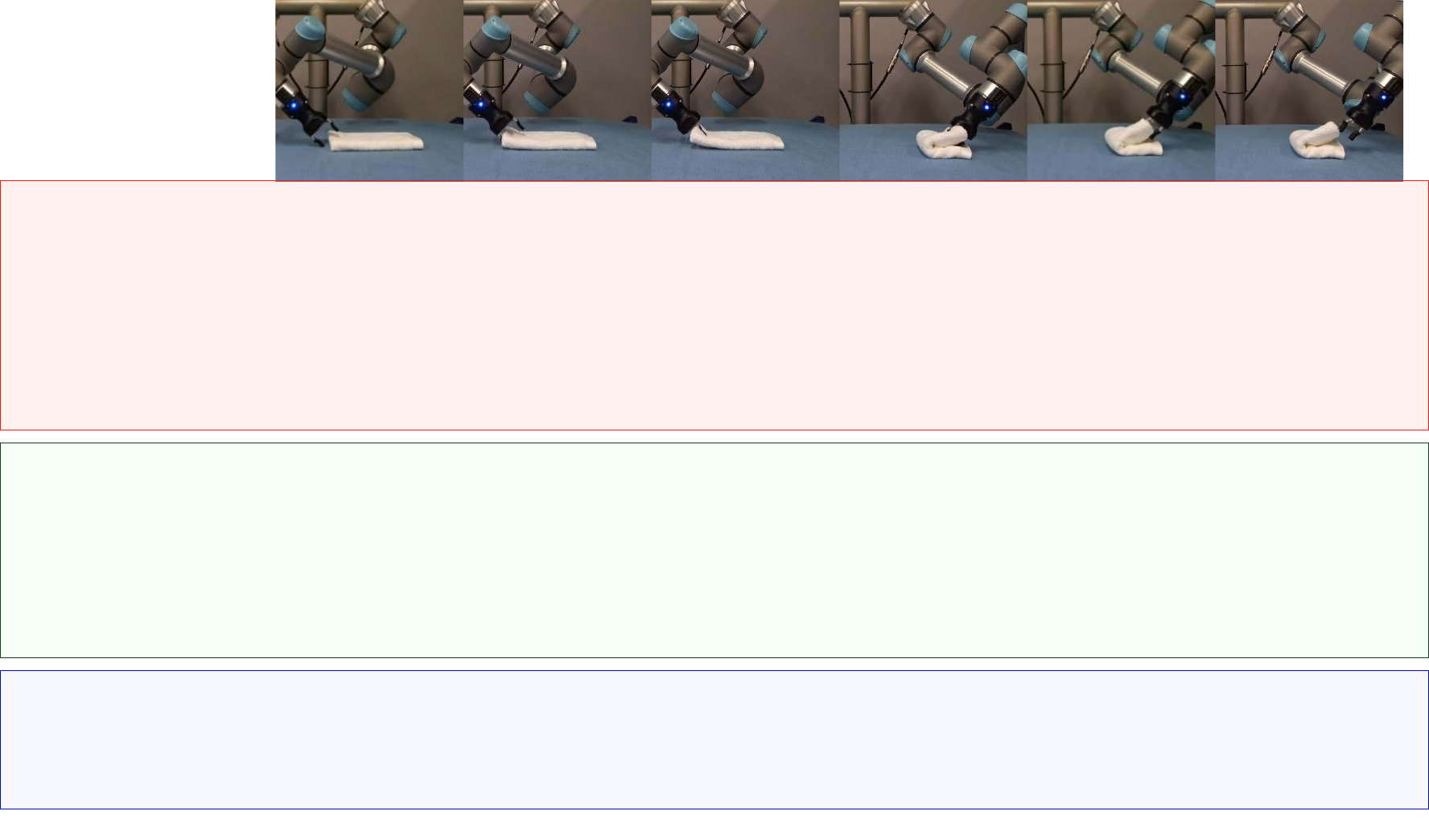
}\\
\subfloat[Task 2: Transport Towel. \ac{taxonomy} no-zero constant category is Struct. Bending (S L2). \label{fig:task-transitions-2}]{
    \def\svgwidth{0.8\linewidth}
    \fontsize{7}{7}
    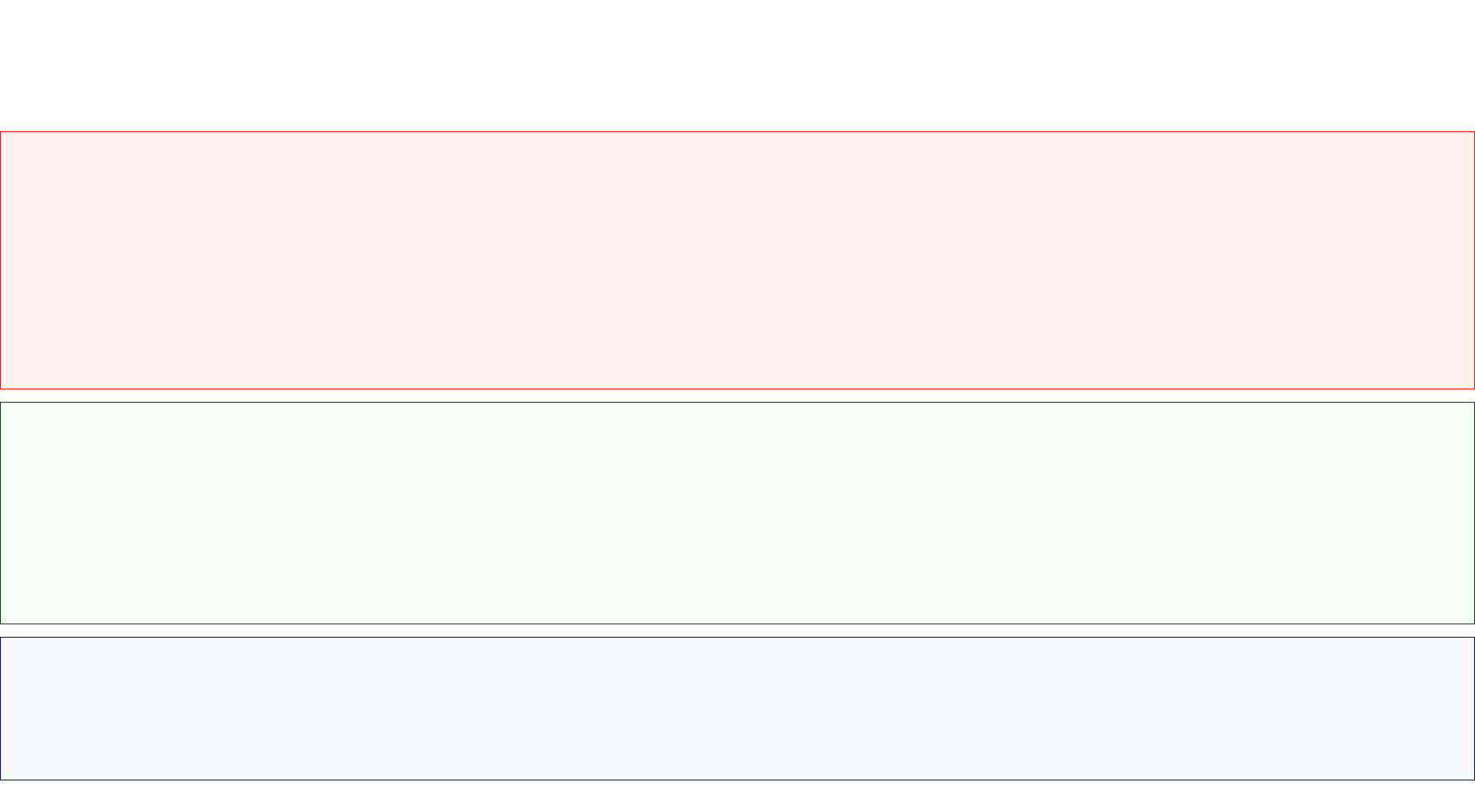
}\\
\subfloat[Task 4: Edge tracing. \ac{taxonomy} no-zero categories are left and right grasp (P and P) and NP Env (R). \cite{paulius_2020_manipulation_tax_embedding} categories are omitted since all are constant during the manipulation actions. \label{fig:task-transitions-4}]{
    \def\svgwidth{0.8\linewidth}
    \fontsize{7}{7}
    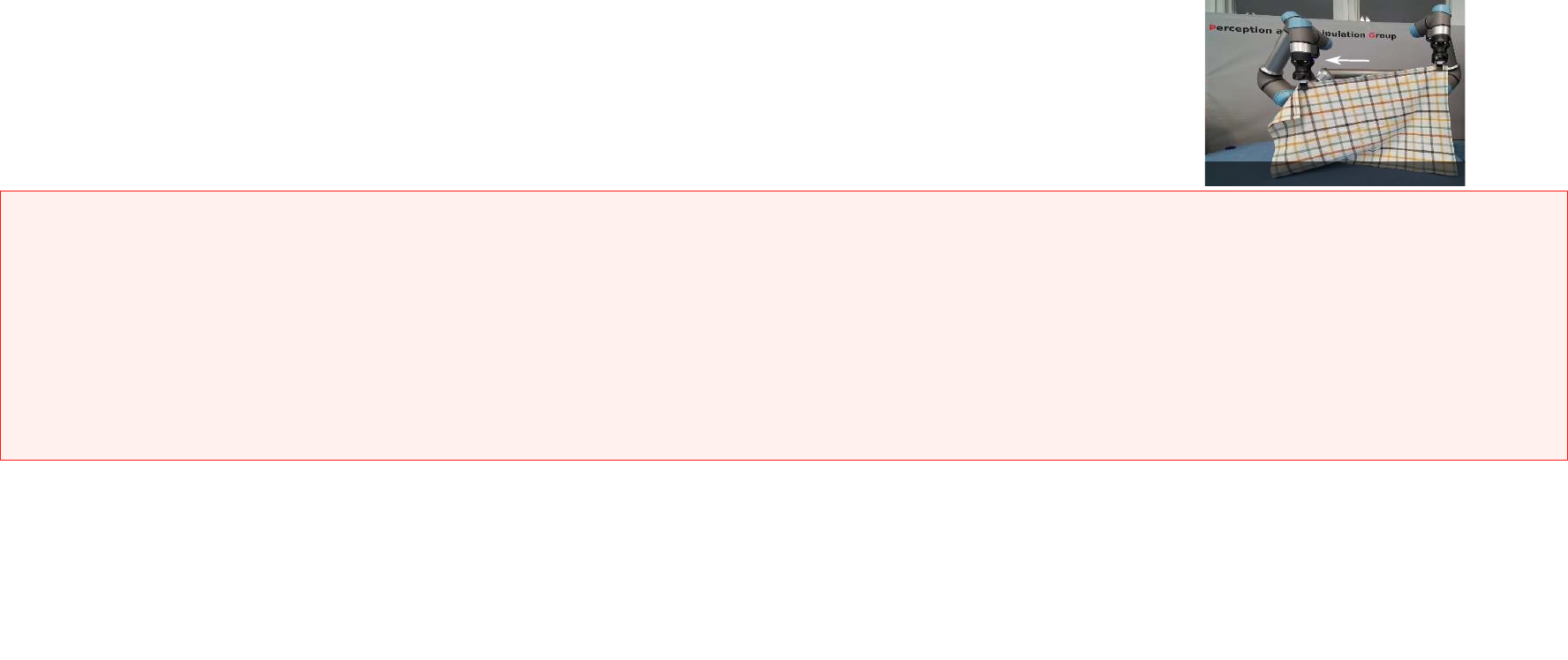
}\\
\caption{Analysis of the transitions of Task 1, Task 2, and Task 4 using the proposed taxonomy, \ac{taxonomy}, as well as \cite{paulius_2020_manipulation_tax_embedding} and \cite{bullock2012hand} taxonomies, showing only the categories that show any change.
For all tasks the structural outcome according to ~\cite{paulius_2020_manipulation_tax_embedding} taxonomy is deforming temporarily, and therefore it is omitted for simplicity.
}
\label{fig:task-transitions}
\end{figure*}

\section{Analysis of the Taxonomy}\label{sec:analysis}
This section presents how the proposed taxonomy can be applied to classify deformable object manipulation tasks, demonstrating the importance of classifying the deformation for understanding the task.
The section starts by introducing ten deformable object manipulation tasks, detailing the robot motions and interactions with objects.
Next, it outlines the methodology used to label the manipulation actions within these tasks according to the proposed taxonomy.
Finally, it describes the experiments performed to analyse the taxonomy, which evaluate the impact of incorporating deformation as a key category alongside the other categories.

\subsection{Deformable Object Manipulation Dataset}
To consider all the interactions of the deformable object with the robot and its surroundings, as well as the motions of the robot, we take inspiration on~\cite{bullock2012hand} and analyse the transitions during robot manipulation tasks.
For such analysis, all the steps that take place during the manipulation of a deformable object need to be recorded. 
Despite the availability of open-source datasets such as the Daily Interactive Manipulation dataset proposed by~\cite{huang_2019_dim_dataset}, current datasets either show only one stage of the manipulation, such as the action of lifting or transporting an object, or have limited deformable object manipulation tasks \citep{mitash_2023_armbench, khazatsky2024droid}.
Thus, we recorded ten deformable object manipulation tasks, five of them using the UR5 robot, and five of them using hand-held grippers. The tasks are divided into five bimanual manipulation tasks and five unimanual tasks.
We recorded RGB-D data of task execution using a Microsoft Azure Kinect.
For the analysis of the proposed taxonomy, we utilised RGB images. 
In order to enable further research on deformable object manipulation we open-source the dataset\footnote{Project website: \scriptsize{\url{https://sites.google.com/view/t-dom}}} including the depth information, which can be used in future works to analyse the deformation of the manipulated objects.
The analysed tasks can be described as follows. 

\paragraph{Task 1: Fold towel.}
The robot grasps a folded towel by sliding under its side to clamp it.
It then performs a quasi-static motion to fold the towel in half before releasing it.

\paragraph{Task 2: Transport towel.}
The robot approaches a folded towel and slides beneath it to clamp it, as in task 1. It then lifts and transports the towel, placing it onto a stack of towels.

\paragraph{Task 3: Wring out towel.}
Two robots approach a towel and slide underneath to grasp it. Then, they lift it from a flat configuration to apply a twisting motion to wring out water.

\paragraph{Task 4: Cloth Edge Tracing.}
In the starting configuration, one robot grasps a corner of a cloth while a second one holds the cloth nearby the edge.
First, the robot grasping the corner moves away from the other manipulator, inducing passive sliding at the edge grasped.
Once the motion stops, the robot holding the edge moves away, creating active sliding along the edge while the first robot remains stationary. 

\paragraph{Task 5: Transport meat.}
The robot approaches a meat-like silicone piece, slides underneath to clamp it securely, and then lifts and transports the piece to place it on a tray.

\paragraph{Task 6: Cloth flattening.}
In this task, a hand-held gripper approaches a cloth with two visible corners. The gripper grasps the cloth, lifts it and places it. Finally, the gripper slides out of the cloth, leaving three visible corners.

\paragraph{Task 7: Unfold medical gown.}
Two hand-held grippers hold a partially unfolded medical gown and perform a dynamic shaking motion to unfold the gown.

\paragraph{Task 8: Bag opening and item insertion}
Similar to task 7, two hand-held grippers holding a bag perform a dynamic motion to improve its opening state. This is followed by the insertion of a rigid object into the bag.

\begin{table*}[h]
\small\sf\centering
\caption{Taxonomy tags describing the ten manipulation tasks and the associated transitions using \ac{taxonomy} according to Figure~\ref{fig:taxonomy}. The tags are divided into: left and right motion (Motion), left and right prehensile grasp (Grasp), non-prehensile environment interaction (N-P. Env.), left and right non-prehensile active contact interaction (N-P. Act.), environment, left and right contact sliding (CS), deformation (Def.), structured bending (S) and unstructured bending (US).
\label{table:taxonomy_codes}}
\begin{tabular}{l l|cccccccc}
\toprule
\multirow{2}{*}{Task} & \multirow{2}{*}{Action}  & \multirow{2}{*}{Motion} & \multirow{2}{*}{Grasp} & \multicolumn{2}{c}{N-P.} & \multirow{2}{*}{CS} & \multirow{2}{*}{Def.} & \multirow{2}{*}{S} & \multirow{2}{*}{US} \\
 & &  &  & Env. & Act. &  &  &  &  \\
\midrule
& 1-1: approach & G N & N N & R & N N & N N N & N & L1 & N \\ 
\rcol & 1-2: slide(under) & G N & N N & R & R N & N A N & N & L1 & N \\ 
 & 1-3: grasp & N N & P N & R & N N & N N N & C & L1 & N \\ 
\rcol & 1-4: fold & GE N & P N & R & N N & N N N & C & L2 & N \\ 
 & 1-5: ungrasp & N N & N N & R & R N & N N N & N & L2 & N \\ 
\rcol \multirow{-6}{*}{Fold towel}  & 1-6: slide(out) & G N & N N & R & R N & N A N & N & L2 & N \\ 
 \hline 
 & 2-1: approach & G N & N N & R & N N & N N N & N & L2 & N \\ 
\rcol & 2-2: slide(under) & G N & N N & R & R N & N A N & N & L2 & N \\ 
 & 2-3: grasp & N N & P N & R & N N & N N N & C & L2 & N \\ 
\rcol & 2-4: transport & G N & P N & N & N N & N N N & C & L2 & N \\ 
 & 2-5: place & GE N & P N & S & N N & N N N & C & L2 & N \\ 
\rcol & 2-6: ungrasp & N N & N N & S & R N & N N N & N & L2 & N \\ 
 \multirow{-7}{*}{Transport towel}  & 2-7: slide(out) & G N & N N & S & R N & N A N & N & L2 & N \\ 
 \hline 
\rcol & 3-1: approach(dual) & G G & N N & R & N N & N N N & N & N & N \\ 
 & 3-2: slide(under-dual) & G G & N N & R & R R & N A A & N & N & N \\ 
\rcol & 3-3: grasp(dual) & N N & P P & R & N N & N N N & N & N & N \\ 
 & 3-4: lift(dual) & G G & P P & N & N N & N N N & TN & L0 & N \\ 
\rcol \multirow{-5}{*}{Wring out towel}  & 3-5: twist(dual) & GE GE & P P & N & N N & N N N & TN+TR & N & N \\ 
 \hline 
 & 4-1: grasp(dual) & N N & P P & R & N N & N N N & N & N & L2 \\ 
\rcol & 4-2: tracing(static) & GE N & P P & R & N N & N N P & TN & L0 & L2 \\ 
 \multirow{-3}{*}{Edge tracing}  & 4-3: tracing(motion) & N GE & P P & R & N N & N N A & TN & L0 & L1 \\ 
 \hline 
\rcol & 5-1: approach & G N & N N & R & N N & N N N & N & N & N \\ 
 & 5-2: slide(under) & G N & N N & R & R N & N A N & N & N & N \\ 
\rcol & 5-3: grasp & N N & P N & R & N N & N N N & C & N & N \\ 
 & 5-4: transport & G N & P N & N & N N & N N N & C & L0 & N \\ 
\rcol & 5-5: place & G N & P N & R & N N & N N N & C & N & N \\ 
 & 5-6: ungrasp & N N & N N & R & R N & N N N & N & N & N \\ 
\rcol \multirow{-7}{*}{Transport meat}  & 5-7: slide(out) & N N & N N & R & R N & N A N & N & N & N \\ 
 \hline 
 & 6-1: approach & G N & N N & R & N N & N N N & N & N & L2 \\ 
\rcol & 6-2: grasp & N N & P N & R & N N & N N N & N & N & L2 \\ 
 & 6-3: lift & G N & P N & R & N N & N N N & N & N & L2 \\ 
\rcol & 6-4: place & G N & P N & R & N N & N N N & N & L1 & L1 \\ 
 \multirow{-5}{*}{Flatten cloth}  & 6-5: slide(out) & N N & N N & R & R N & N A N & N & L1 & L1 \\ 
 \hline 
\rcol & 7-1: grasp(dual) & N N & P P & N & N N & N N N & N & N & L2 \\ 
 \multirow{-2}{*}{Unfold gown}  & 7-2: unfold(dynamic) & K K & P P & N & N N & N N N & N & N & L1 \\ 
 \hline 
\rcol & 8-1: grasp(dual) & N N & P P & N & N N & N N N & N & N & L0 \\ 
 & 8-2: unfold(dynamic) & K K & P P & N & N N & N N N & N & N & L0 \\ 
\rcol & 8-3: grasp(dual) & N N & P P & N & N N & N N N & N & N & N \\ 
 \multirow{-4}{*}{Bag opening}  & 8-4: offer & N N & P P & R & N N & N N N & N & N & N \\ 
 \hline 
\rcol & 9-1: approach & N G & N N & RS & N N & N N N & N & N & L1 \\ 
 & 9-2: contact & N G & N N & RS & N R & N N N & N & N & L1 \\ 
\rcol & 9-3: grasp & N N & N P & RS & N R & N N N & TN & N & L1 \\ 
 & 9-4: lift & N G & N P & N & N N & N N N & N & N & L1 \\ 
\rcol & 9-5: approach(second) & G N & N P & N & N N & N N N & N & N & L1 \\ 
 & 9-6: grasp(dual) & N N & P P & N & N N & N N N & N & N & L1 \\ 
\rcol & 9-7: ungrasp(second) & N N & P N & N & N N & N N N & N & N & L0 \\ 
 & 9-8: approach(second) & N G & P N & N & N N & N N N & N & N & L0 \\ 
\rcol & 9-9: grasp(dual) & N N & P P & N & N N & N N N & N & N & L0 \\ 
 \multirow{-10}{*}{Open glove}  & 9-10: open & GE GE & P P & N & N N & N N N & TN & L0 & N \\ 
 \hline 
\rcol & 10-1: grasp & N N & N P & N & N N & N N N & N & L0 & N \\ 
 & 10-2: hang & N GE & N P & N & N N & N N N & N & L0 & N \\ 
\rcol & 10-3: grasp(dual) & N N & P P & N & N N & N N N & N & L0 & N \\ 
 & 10-4: ungrasp(second) & N N & P N & N & N N & N N N & N & L0 & N \\ 
\rcol & 10-5: approach(second) & G G & P N & N & N N & N N N & N & L0 & N \\ 
 & 10-6: grasp(dual) & N N & P P & N & N N & N N N & N & L0 & N \\ 
\rcol & 10-7: fold(cable) & GE GE & P P & N & N N & N N N & N & L1 & N \\ 
 & 10-8: ungrasp(hold) & N N & N P & N & R N & N N N & N & L1 & N \\ 
\rcol & 10-9: hang & N GE & N P & N & R N & N N N & N & L1 & N \\ 
 & 10-10: grasp(dual) & N N & P P & N & N N & N N N & N & L1 & N \\ 
\rcol \multirow{-11}{*}{Cable looping} & 10-11: ungrasp(second) & N N & P N & N & N N & N N N & N & L1 & N \\ 
\bottomrule
\end{tabular}\\[10pt]
\end{table*}

\paragraph{Task 9: Open surgical glove.}
One hand-held gripper is inserted into a box to pinch and lift a glove. Once lifted, a second gripper grasps the glove's wrist border.
The first gripper releases the glove and re-grasps it from the opposite side of the wrist. Finally, both grippers move to open the glove. 

\paragraph{Task 10: Cable looping.}
In the starting configuration, a cable is grasped by a hand-held gripper. The gripper moves to hang the cable onto a second gripper, which secures the cable while the first gripper forms a loop. Once the loop is created, the second gripper is opened to hang the loop.

The tasks have been designed to cover at least the basic types of deformation except shearing, which is less common in manipulation tasks. The compression deformation takes place in tasks 1, 2, 5 and 6; tension in tasks 3, 4 and 9; torsion in task 3; and either structured or unstructured bending in all tasks. 
In addition, the tasks have been designed to include commonly studied manipulation tasks, encompassing a diverse range of actions and trying to maximise the variety of manipulation skills required.

\subsection{Task Segmentation and Representation}\label{sec:analysis-task-segmentation}
The tasks described in the previous section are composed of a sequence of sub-tasks or actions, each of which can be described by \ac{taxonomy} tags.
To identify the segmentation induced by \ac{taxonomy}, we manually label the time instants in which a robot action creates a change in any of the taxonomy categories.

The resulting segmentation for Tasks 1, 2 and 4 is shown in Figure~\ref{fig:task-transitions}, while an overview of the segmentation for all the tasks, along with their corresponding tags, is provided in Table~\ref{table:taxonomy_codes}.
To distinguish between the left and right manipulators in bimanual tasks, the tags are written in the left and right side of the category, accordingly. 
For unimanual tasks, we retain both tags but set the second manipulator to a none tag, ensuring that all taxonomy tags have the same structure, regardless of the number of manipulators. 

In order to evaluate the contribution of our proposed taxonomy to the state-of-the-art, we compare against the taxonomies proposed by  \cite{bullock2012hand} and \cite{paulius_2020_manipulation_tax_embedding}.
The Bullock taxonomy was created to represent dexterous manipulation, with special attention to in-hand manipulation. Although it is defined as a decision tree, it can also be described in tags with the following categories:
\begin{itemize}
\setlength\itemsep{0.1cm}
    \item \textbf{Contact Category}: Contact Non-Prehensile (C NP); Contact Prehensile (C P);
     No Contact (NC).
    \item \textbf{Motion Category}:
    Motion within hand (M W); 
    Motion not within hand (M NW); 
    No Motion (NM).
    \item \textbf{Slippage category}:	
        Motion at contact (A);
        No motion at contact (NA);
        No contact ( "-").
\end{itemize}
Note that it is assumed that arm motion and motion within hand will not happen simultaneously.

For the taxonomy of \cite{paulius_2020_manipulation_tax_embedding} the trajectories of the robot or hand-held grippers as well as the manipulated object are classified by the corresponding number of moving axes.
Furthermore, the type of interaction distinguishes between contacts of the manipulator with rigid surfaces (rigid discontinuous), contact with the deformable object (soft discontinuous), and the object that is grasped (soft continuous).
Finally, all the manipulation tasks involve objects working in their elastic region of the stress-strain curve, and therefore, our tasks involve either no deformation or temporary deformation of the object.

To better compare with these taxonomies,  Figure~\ref{fig:task-transitions} also shows the segmentations induced by our \ac{taxonomy}, Bullock's and Paulius, that act as baselines. 
The actions in task 1, shown in Figure~\ref{fig:task-transitions}-a, involve: approaching and releasing the object (Q.Grav), folding the object (Q.Grav/Elast), a continuous contact with the rigid table (N-P.R), and a non-prehensile contact with the gripper (NP.A R) when it slides beneath one side of the cloth to clamp it (CS active).  
In terms of deformation, the cloth is initially folded with a single g-fold (structured bending level 1), grasping the object results in compression deformation within the cloth, and the folding action increases the structured bending level to SL2.
While the segmentation remains the same across all taxonomies, there are several key differences. 
Firstly, passive contacts with the environment are not considered in the baselines. 
For example, the Bullock taxonomy describes the folding action equally as any object transfer, that is, a grasp in motion.
In contrast, \ac{taxonomy} provides a more detailed representation.
In addition to the grasp in motion, it includes a description of the point grasp, the passive contact with the environment, and a change in deformation.   
Secondly, throughout all the manipulation steps \cite{paulius_2020_manipulation_tax_embedding} classifies the deformation invariant, as the object is temporarily deforming.
Therefore, the action of folding does not produce any change in their deformation category.

In the transport task shown in Figure~\ref{fig:task-transitions}-b, the structured bending deformation remains constant due to the folded configuration of the fabric.
However, placing a folded object requires a placing strategy involving an environmental contact between the object and the placing location. This interaction introduces a segment between the actions 4 and 5. 
In particular, this segment is not present in the \cite{bullock2012hand} taxonomy, as it does not account for passive environmental contacts.

The classification differences with \cite{paulius_2020_manipulation_tax_embedding} taxonomy become more pronounced when examining the action of edge tracing in Figure~\ref{fig:task-transitions}-c.
Here, \cite{paulius_2020_manipulation_tax_embedding} classification shows three constant categories, which refer to the motion of both robots which are in constant contact with the cloth, temporally deforming the object. Therefore, Paulius taxonomy provides only one segment. 
Both \ac{taxonomy} and \cite{bullock2012hand} taxonomy are able to distinguish between actions T5-2 tracing (static) and T5-3 tracing (motion), where contact sliding or motion at contact plays a crucial role.
In addition, our classification enables us to distinguish the change in the structure of the cloth which starts unstructured (US L1) and transitions to structured bending deformation (SL0) once the task is complete.

Overall, the classification using \ac{taxonomy} provides similar granularity to \cite{bullock2012hand} when segmenting the tasks, except in the tasks involving passive contacts, like placing an object.
Here, a first contact with the environment is used to place the deformable object, like in task 2.
More importantly, the classification using \cite{paulius_2020_manipulation_tax_embedding} and \cite{bullock2012hand} taxonomies provide less understanding of the transitions the robot needs to accomplish the task since the deformation phases are disregarded in the classification.

\begin{figure*}[t]
\centering
\def\svgwidth{\linewidth}
{
    \fontsize{5}{5}
    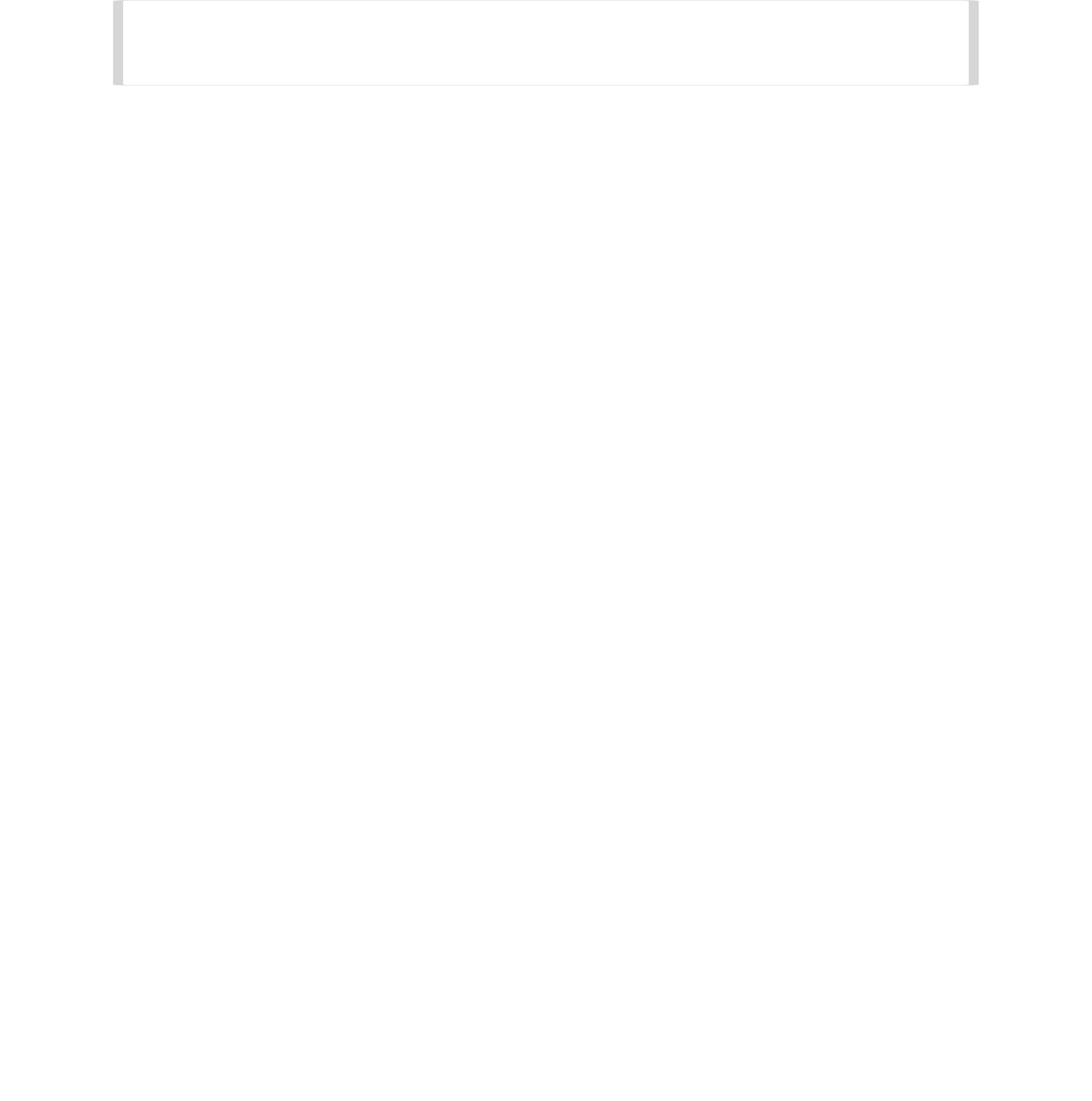}
    \caption{Graph clustering of manipulation actions in the proposed dataset for a) \ac{taxonomy}, b) \ac{taxonomy} without deformation, c) \cite{paulius_2020_manipulation_tax_embedding} taxonomy, and d) \cite{bullock2012hand} taxonomy.
    The nodes represent the action-IDs and the edges action-IDs which share the same tag according to the evaluated taxonomy. The colours of the nodes represent the deformation of the object. The edges and clusters colours indicate node connectivity. The markers represent no bending (circle), structured bending (triangle up), unstructured bending (triangle down), and both structured and unstructured bending (square).
    Note that all graphs are complete graphs but only one edge is shown for clarity.}
    \label{fig:graph_full}
\end{figure*}

\subsection{Evaluation of the Taxonomy}
This section provides an analysis of the capability a taxonomy has to classify the entire set of actions depicted in Table~\ref{table:taxonomy_codes}.
Given the diverse range of deformations within the action set, taxonomies that group together manipulation skills producing different deformations may be considered less suitable for classifying deformable object manipulation tasks.

To perform this analysis, we first define action-IDs like the ordered set of tags that describe each action, as appearing in Table \ref{table:taxonomy_codes}. We then create a graph where the nodes correspond to an action-ID and an edge connects two nodes when they have the same code. Figures~\ref{fig:graph_full}-a, \ref{fig:graph_full}-b, \ref{fig:graph_full}-c, and \ref{fig:graph_full}-d, show clusters of action-ID graphs for \ac{taxonomy}, \ac{taxonomy} without the deformation category, \cite{paulius_2020_manipulation_tax_embedding} taxonomy, and \cite{bullock2012hand} taxonomy, respectively.
Furthermore, each node shape and colour is associated with a deformation as classified in Table~\ref{table:taxonomy_codes}, which helps to identify different deformations that are classified together.
The node colours serve to differentiate the combinations of deformations present, while each node shape represents a type of bending deformation: circles for no bending, upward triangles for structured bending, downward triangles for unstructured bending, and squares for both types.

The cluster results in Figure~\ref{fig:graph_full}-a show that our taxonomy can clearly separate actions that generate different deformations, highlighting its capability to categorise disjoint actions by the object deformation involved.
This classification results in only four clusters with more than one action-ID, each sharing motion type, interactions and object deformation tags.
For instance, the edge between action-IDs "T1-6 slide (out)" and "T2-2 slide (under)" originates from the same structured bending deformation of the towel in task 2 (transport towel) as the end of task 1 (fold towel).

To assess the role of the deformation category in distinguishing between actions, we conducted an ablation study by removing the deformation category and re-clustering the action-IDs, as shown in Figure~\ref{fig:graph_full}-b.
Without the deformation category, a large cluster of eight "grasp (dual)" actions appears.
Additionally, two smaller clusters appear involving "grasp" and "grasp (dual)" actions.
This indicates that the grasp actions are the same except for the deformation of the grasped object.
This deformation is relevant for decision making in tasks like grasping a folded object, where the control strategy needs to retain the object state. 
In particular, some grasps are separated due to the non-prehensile environment interactions.
For instance, the "T9-3: grasp" action is a pinch grasp where the grasped object (glove) is in contact with a rigid object (box) and soft object (glove underneath).
Similarly, the "T10-1: grasp" action is performed in the air, that is, without an environment contact with the table as seen in other grasps across the dataset.
The impact of the structured and unstructured bending is also clear: actions exhibiting different bending deformations are grouped together, such as the slide  actions T1-2, T1-6, T2-2 and T5-2, or the approach actions T1-1, T2-1, T5-1 and T6-1.
As previously discussed, such distinction is relevant for the control strategy. For example, ignoring the deformation when sliding under a folded object might lead to undesired unfolding.
Overall, these results highlight the importance of the deformation information to classify semantically the manipulation actions.

Next, we evaluated the taxonomy proposed by \cite{paulius_2020_manipulation_tax_embedding}, presented in Figure~\ref{fig:graph_full}-c.
This taxonomy produces 22 clusters, with the largest containing 17 action-IDs, primarily from diverse grasping actions, where quite varied deformations are grouped together. 
Although this taxonomy accounts for deformable objects, it simplifies deformation into the categories of temporal or permanent deformation.
This approach is effective for classifying actions that result in a permanent deformation, but does not explicitly address manipulation skills that operate within an object's stress-strain elastic region.
Consequently, actions that compress, stretch, twist or bend an object might be grouped together given the motion and interactions with the environment that generate the deformation are the same.
This reflects the objective of the taxonomy, which was to classify manipulation from a robot motion perspective, rather than classifying the different types of deformation that take place throughout a manipulation task, as in \ac{taxonomy}.
In addition, the taxonomy does not explicitly classify prehensile grasps,which limits its ability to distinguish manipulations that involve different contact points.
Altogether, this makes the taxonomy less suited for classifying the large repertoire of skills required for the robotic manipulation of deformable objects.

Finally, we examine the results from the taxonomy proposed by \cite{bullock2012hand}, shown in Figure~\ref{fig:graph_full}-d. 
This taxonomy creates a total of 21 clusters which, although lower than those using \cite{paulius_2020_manipulation_tax_embedding} taxonomy, it provides more granularity by classifying prehensile and non-prehensile motions, contacts within the hand, and motion at contact.
This is reflected by a smaller largest cluster of 10 action-IDs (compared to the 17 action-IDs in Figure~\ref{fig:graph_full}-c).
This cluster of "grasp" actions resembles that appearing in our taxonomy without deformation in Figure~\ref{fig:graph_full}-b.
Similarly, actions such as lifting, transporting, and folding are grouped together because deformation is not considered.
As previously mentioned, the deformation information is relevant because it imposes constraints to the possible motions, and therefore, the planning of actions. 
In addition, Bullock's taxonomy does not consider environment contacts, which are important in tasks like placing.
This omission explains why placing actions are grouped together with transport, lifts and folds in Bullock's taxonomy, while they are separated in Figure~\ref{fig:graph_full}-b.
These passive contacts with the environment are relevant in rigid manipulation to generate external forces, which are implicitly addressed in Bullock's taxonomy in categories 5 and 7, where passive motion at contact occurs due to an external force.
However, these contacts are even more crucial for deformable objects such as clothes, where they act as a third agent that shapes the geometry of the object \citep{borras_2020_grasp_cloth_analysis}.
Hence, by adding the category \textit{non-prehensile environment} we can treat the environment as an additional gripper within the system, which is relevant in both high- and low-level planning

It is important to note that through our analysis we are not suggesting that the baselines taxonomies are incorrect.
These taxonomies were developed with different objectives in mind.
Our aim is to point that these are missing semantic tags relevant for the robotic manipulation of deformable objects which are incorporated in our proposed taxonomy.
\section{Discussion \& Future Work}\label{sec:discussion}
This section discusses future directions as well as potential applications of the taxonomy to key aspects of deformable object manipulation.

\subsection{Extending Deformation Categories}\label{sec:discussion-def-categories}
The proposed taxonomy provides an understandable categorisation of the deformations that can arise when manipulating deformable objects.
However, other types of deformations, such as tearing and cutting \citep{, heiden2021disect} or permanent deformation of an object~\citep{shi_2024_robocraft_gns_ijjr} are not incorporated into \ac{taxonomy}.
These deformations involve the irreversible breaking of material bonds, making them significantly more difficult to predict and control compared to deformations like bending or stretching.
The study of such permanent deformations is in its early stages, but there are ongoing efforts to address these complex deformation types~\citep{xu2023roboninja, Haiderbhai_2024_cutting}.

Future advances in deformable manipulation tasks will allow the taxonomy to integrate further deformation types, enhancing its applicability to a broader range of tasks.

\subsection{Measuring Deformation}\label{sec:discussion-measuring-deformation}
The presented categories for deformation have been given different levels of attention in literature depending on the type of objects. For volumetric, mostly compression is studied. In contrast, for clothes or cables most attention is put on bending.
Indeed,
there are several efforts to accurately measure the bending deformation of an object~\cite{sanchez_2018_manipulation_sensing_do_survey}, which is critical in the precise manipulation of deformable objects~\cite{longhini2024unfolding}.
One example of these efforts is the derivative of the Gauss Linking Integral~\cite{coltraro2023representation}, which can provide a low-dim topological representation of cloth state that allows the classification of different bending classes only by distance.
While other types of deformation such as tension and shear have been less extensively explored in cloth manipulation, these are highly relevant for planning the control strategies.

At the moment, there is no consensus regarding the sensing and representation of deformation.
Our taxonomy aims to pave the way of categorising deformable object manipulation, which could be integrated with techniques that can simultaneously measure deformation and force to plan manipulation. 

\subsection{Assisted Gripper Design}
The process of gripper design requires expertise in multiple domains, as well as significant efforts involving the refinement of the design~\citep{babin_2021_review_gripper_design}. 
Manipulation taxonomies such as \cite{bullock2012hand} have influenced several gripper designs \citep{rojas2016gr2, zhou2018bcl}.
These taxonomies can assist by providing a structured framework to analyse the gripper manipulation modes.
As an example, this can enable engineers to tailor specific features that effectively perform manipulation tasks such as in-hand manipulation~\citep{xie_2024_inhand_parallel_jaw_bullock}.
Here, \ac{taxonomy} can further aid researchers to devise grippers that specialise in deformable object manipulation tasks, such as grippers designed for textile manipulation~\citep{hinwood_2020_textile_gripper_design}.

To overcome the arduous task of hand-engineering gripper designs for a specific task, co-design algorithms emerged as an alternative to alleviate such human efforts~\citep{demiel_2017_codesign_gripper, chen2020hardware, kim2021mo, dong2024co}.
These approaches have largely focused on rigid object manipulation, leaving designs for deformable object manipulation under-explored due to the complex nature of deformable objects and their interactions.
\ac{taxonomy} offers a structured framework to potentially bridge this gap, facilitating robot design tailored to deformable objects. 
For instance, retrieving a pancake from a pan highlights the importance of deformation constraints, where in order to preserve the pancake's integrity prehensile grasps need to be avoided, as this would risk destroying the object. 
Instead, strategies such as contact sliding beneath the pancake and non-prehensile interactions over a wide contact area are preferred~\citep{beetz_2011_pancakes_robot}.

By systematically applying \ac{taxonomy} subcategories to define task-specific criteria, either for hand-engineered or co-design algorithms, the morphology of a gripper can be adapted to achieve the desired manipulation goals.
\subsection{Towards General Manipulation Skills}\label{sec:discussion-general-manipulation}
Finally, in deformable object manipulation, policies are often designed for specific tasks~\citep{zhang_2022_assisted_dressing_science, driess_2023_nerf_gnn_rope, tran_2023_sponge}, limiting their broader applicability. 
However, many manipulation actions are shared across different tasks.
As shown in our experiments, \ac{taxonomy} can identify these shared actions, enhancing policy generalisation.
For instance, tasks with the same taxonomy code may involve similar actions, allowing a policy trained on one task to be adapted to another. 
This identification can improve the efficiency and versatility of robotic systems, reduce the need for retraining, and enable quicker adaptation to new tasks.

Despite the recent advances in language models for manipulating both rigid and deformable objects~\citep{openx_2024_icra, kim24openvla}, most of these approaches neglect the deformation information of the object.
In this context, the identification of shared actions and different deformations using \ac{taxonomy} could facilitate learning multi-task policies for complex deformable object manipulation tasks.

\section{Conclusion}
This work introduced the \acf{taxonomy}, aimed to classify the skills required to perform robotic manipulation of deformable objects. The taxonomy focuses on three crucial components: robot motion, the interactions with the robot end-effector or the environment, and most importantly, object deformation.
Key to this taxonomy is a deformation classification that draws from mechanical engineering concepts to categorise deformation via compression, tension, bending, torsion, and shear.
Here, we discussed the challenges of identifying different bending deformations, and, to circumvent such challenge, proposed qualitative metrics to classify the existence of complex creasing as structured and unstructured bending, respectively. 
Then, we introduced a classification for robot motion from a dynamics and energy perspective, integrating quasi-static and dynamic manipulation. 
Next, for interaction types, we defined prehensile grasps based on the contact constraint geometry, non-prehensile interactions to distinguish between the environment and agent interactions, and included the special case of contact sliding, also referred to as motion at contact. 

To evaluate the taxonomy's effectiveness in characterising deformable object manipulation, we curated a dataset of tasks involving various deformable objects such as cloths, meat-like silicone, and surgical aprons and gloves. 
We applied \ac{taxonomy} to classify the task manipulation transitions, representing the actions as action-IDs inspired by prior taxonomies.
Our analysis demonstrated that classifying the type of deformation is essential for distinguishing similar robot actions across different deformable objects, where the proposed taxonomy offers a comprehensive framework for classifying the manipulation strategies required for these tasks.

Finally, we discussed potential applications and research directions of the proposed taxonomy, connecting the discussed types of deformation with methods for accurately measuring deformation, and providing a generic view into general manipulation policies.
All in all, we believe \ac{taxonomy} lays a solid foundation for advancing the understanding and development of deformable object manipulation.



\begin{acks}
The authors would like to thank the technical staff from IRI, CSIC-UPC Pablo Salido Luis-Ravelo for their support to set-up the recording of the manipulation dataset.
\end{acks}

\begin{dci}
The author(s) declared no potential conflicts of interest with regard to the research, authorship, and/or publication of this article.
\end{dci}

\begin{funding}
This work was supported by the European Union’s Horizon Europe Programme through the project SoftEnable (HORIZON-CL4-2021-DIGITAL-EMERGING-01-101070600) and the MCIN/ AEI /10.13039/501100011033 through the project PID2020-118649RB-I00 (CHLOE-GRAPH).
\end{funding}

\bibliographystyle{SageH}
\bibliography{literature,simulation_and_animation,surveys_and_books, deform}

\end{document}